\documentclass{article}
\usepackage{spconf,amsmath,epsfig}

\usepackage{amsfonts}
\usepackage[colorlinks]{hyperref}
\usepackage{multirow}
\usepackage{tabularx}

\usepackage{amssymb}
\usepackage{pifont}
\begin{document}\sloppy

\def\x{{\mathbf x}}
\def\L{{\cal L}}

\title{Ensembles of feedforward-designed convolutional neural networks}

\name{Yueru Chen, Yijing Yang, Wei Wang and C.-C. Jay Kuo}
\address{University of Southern California, Los Angeles, California, USA}

\maketitle
\ninept
\begin{abstract}

An ensemble method that fuses the output decision vectors of multiple
feedforward-designed convolutional neural networks (FF-CNNs) to solve
the image classification problem is proposed in this work. To enhance
the performance of the ensemble system, it is critical to increasing the
diversity of FF-CNN models. To achieve this objective, we introduce
diversities by adopting three strategies: 1) different parameter
settings in convolutional layers, 2) flexible feature subsets fed into
the Fully-connected (FC) layers, and 3) multiple image embeddings of the
same input source. Furthermore, we partition input samples into easy and
hard ones based on their decision confidence scores. As a result, we can
develop a new ensemble system tailored to hard samples to further boost
classification accuracy.  Experiments are conducted on the MNIST and
CIFAR-10 datasets to demonstrate the effectiveness of the ensemble method.

\end{abstract}
\section{Introduction}\label{sec:intro}

We have seen rapid developments in the literature of convolutional
neural network (CNN) in last six years
\cite{Lecun98gradient-basedlearning, krizhevsky2012imagenet,
He_2016_CVPR, huang2017densely}. The CNN technology provides
state-of-the-art solutions to many image processing and computer vision
problems. Given a CNN architecture, all of its parameters are determined
by the stochastic gradient descent (SGD) algorithm through
backpropagation (BP). The BP training demands a high computational cost.
Furthermore, most CNN publications are application-oriented. There is a
limited amount of progress after the classical result in
\cite{cybenko1989approximation}. Examples include: explainable CNNs
\cite{zhang2017interpretable, kuo2016understanding, kuo2017cnn} and
feedforward designs without backpropagation \cite{chen2017saak,
kuo2018data, kuo2018interpretable}. 

The determination of CNN model parameters in the one-pass feedforward
(FF) manner was recently proposed by Kuo {\em et al.} in
\cite{kuo2018interpretable}. It derives network parameters of a target
layer based on statistics of output data from its previous layer.  No BP
is used at all.  This feedforward design provides valuable insights into
the CNN operational mechanism. Besides, under the same network
architecture, its training complexity is significantly lower than that
of the BP-design CNN. FF-designed and BP-designed CNNs are denoted by
FF-CNNs and BP-CNNs, respectively. 

The FF-CNN and the BP-CNN were applied to the MNIST
\cite{lecun1998gradient} and CIFAR-10 \cite{krizhevsky2009learning}
datasets for performance benchmarking in \cite{kuo2018interpretable}.
The BP-CNN outperforms the FF-CNN by a small margin in terms of
classification accuracy.  To improve the performance of the FF-CNN, we
use multiple FF-CNNs as base classifiers in an ensemble system and show
that the ensemble idea offers a promising solution to reach higher
classification accuracy in this work.  Although the ensemble idea can be
applied to both BP-CNNs and FF-CNNs, it is more suitable for FF-CNNs
since FF-CNNs are weaker classifiers of extremely low complexity. We
conduct an extensive performance study on the BP-CNN and the ensemble of
FF-CNNs against the MNIST and CIFAR-10 datasets.  Besides, we report the
results by splitting simple and hard examples and treating them
separately. 

\begin{figure*}[htb]
\centering
\includegraphics[width=14cm]{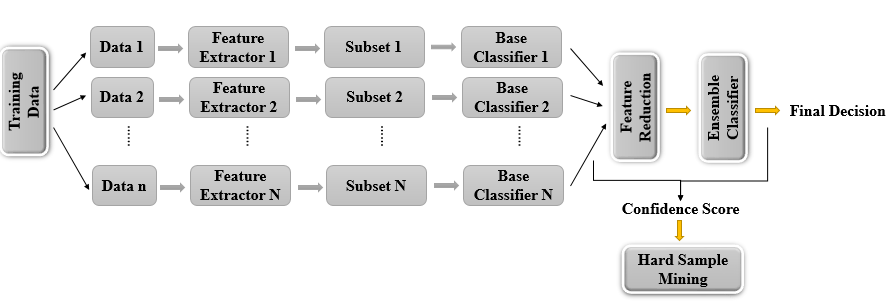}
\caption{Overview of the proposed FF-CNN ensemble method.}\label{fig:overview}
\end{figure*}

This work has several novel contributions. First, we make one
modification on the FF-CNN design to achieve higher classification
performance. That is, we apply the channel-wise PCA to spatial outputs
of the conv layers to remove spatial-dimension redundancy. This reduces
the dimension of feature vectors furthermore.  This will be elaborated
in Sec. \ref{subsec:FF-CNN}. Second, our major contribution is to
develop various ensemble systems using multiple FF-CNNs as base
classifiers. The idea is shown in Fig.  \ref{fig:overview}. To boost the
performance of the ensemble solutions, we introduce three diversities:
1) flexible parameter settings in conv layers, 2) subsets of derived
features, and 3) flexible image input forms.  Third, we define the
confidence score based on the final decision vector of the ensemble
classifier and use it to separate easy examples from hard ones.  Then,
we can handle easy and hard examples separately. 

The rest of this work is organized as follows. The background material
is reviewed in Sec. \ref{sec:review}. The proposed methods are presented
in Sec.  \ref{sec:proposed}. Experimental results are given in
Sec.\ref{sec:experiment}. Finally, concluding remarks are drawn in Sec.
\ref{sec:conclusion}. 

\section{Background}\label{sec:review}

{\bf FF-CNN.} An FF-CNN consists of two modules in cascade: 1) the
module of convolutional (conv) layers and 2) the module of
fully-connected (FC) layers. They are designed using completely
different strategies. 

The construction of conv layers is totally unsupervised since no labels
are needed in the construction process.  It is designed in
\cite{kuo2018interpretable} as subspace approximation via spectral
decomposition followed by the maximum pooling in the spatial domain.
The subspace approximation is obtained using a new signal transform
called the Saab ({\bf S}ubspace {\bf a}pproximation with {\bf a}djusted
{\bf b}ias) transform. The Saab transform is a variant of the principal
component analysis (PCA). It has a default constant-element bias vector
used to annihilate nonlinear activation. Each conv layer contains one
Saab transform followed by a maximum spatial pooling operation.  The
maximum pooling operation, which plays the "winner-takes-all" role, is a
nonlinear one.  Its model parameters of a target layer are derived from
the statistics of the output of the previous layer. The feature
discriminant power is increased gradually due to a larger receptive
field. 

The design of FC layers casts as a multi-stage linear least squared
regression (LSR) problem in \cite{kuo2018interpretable}.  Suppose that
the input and output dimensions are $K_i$ and $K_o$ ($K_i > K_o$),
respectively. We can cluster training samples of dimension $K_i$ into
$K_o$ clusters and map all samples in a cluster into the unit vector of
a vector space of dimension $K_o$.  Such a unit vector is also known as
the one-hot vector. The index of the output space dimension defines a
pseudo-label. 

\noindent
{\bf Ensemble Methods.} Ensembles are often used to integrate multiple
weak classifiers and make them be a stronger one
\cite{zhang2012ensemble}.  Examples include bagging
\cite{breiman1996bagging}, the random forest (an ensemble of decision
trees) \cite{breiman2001random}, stacked generalization
\cite{wolpert1992stacked}, etc.  
Ensemble methods may not necessarily result in
better classification performance than individual ones.  Diversity is
critical to the success of an ensemble system \cite{brown2005diversity,
kuncheva2003measures}. 

\begin{figure}[h!]
\centering
\includegraphics[width=7cm]{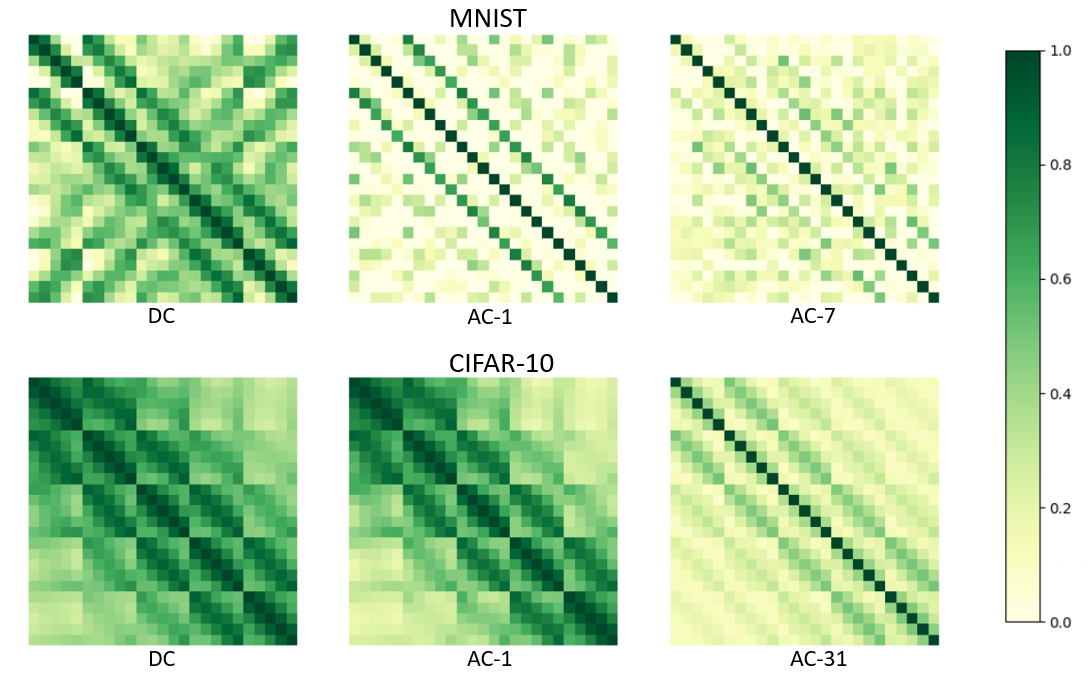}
\caption{Visualization of correlation matrices of DC and AC filter
responses in the last conv layer of the LeNet-5-like FF-CNNs against the MNIST and
CIFAR-10 datasets.} \label{fig:corr}
\end{figure}

\section{Proposed Method}\label{sec:proposed}

We use multiple FF-CNNs to serve as baseline classifiers to construct an
ensemble system. Several novel ideas are proposed to make the ensemble
system more effective. 

\subsection{Channel-wise PCA (C-PCA) for Spatial Dimension Reduction}
\label{subsec:FF-CNN}

Although the Saab transform can reduce redundancy in the spectral
domain, there still exists correlation among spatial dimensions of the
same spectral component.  This is illustrated in Fig.  \ref{fig:corr}.
We see that the correlation is stronger in low frequency components.
Also, by comparing the MNIST and CIFAR-10 datasets, the correlation is
stronger in the CIFAR-10 dataset.  To further reduce the feature space
dimension, we apply the PCA to spatial dimensions at each filter. This
is called the channel-wise PCA (C-PCA). 

We use indices $l$ and $k$ to denote the $l$th conv layer and the $k$th
spectral component. The feature dimension after the Saab transform is
$S_l \times W_l \times H_l$, where $S_l$, $W_l$ and $H_l$ are the
spectral, width and height dimensions of the $l$th conv layer. Then, we
apply C-PCA to features of the same filter index, $k$, and reduce the
original dimension, ${W_k \times H_k}$, to a space of smaller dimension
$L_k$, where $L_k < (W_k \times H_k)$, Thus, the feature dimension of a
certain conv layer is equal to $\sum_{k=0}^{M-1} L_k$, where is $M$ is
the selected filter number. 

\subsection{Diversity}\label{sec:methods}

To improve the performance of an FF-CNN baseline, we develop a simple
yet effective ensemble method as illustrated in Fig.
\ref{fig:overview}. Here, we consider ensembles of LeNet-like CNNs,
which contain two convolutional layers, two FC layers and one output
layer. We adopt multiple FF-CNNs as the first-stage base classifiers in
an ensemble system and concatenate their output decision vectors, whose
dimension is the same as the class number. Then, we apply PCA to reduce
feature dimension before feeding them into the second-stage ensemble
classifier.  The success of ensemble systems highly depends on the
diversity of base classifiers. We propose three ways to increase the
diversity of the baseline FF-CNNs as elaborated below. 

\noindent
{\bf Scheme 1) Flexible parameter settings in conv layers.} We choose different
filter sizes in conv layers.  The filter spatial dimension is the same
for the two conv layers of the LeNet-5 (i.e. $(5 \times 5, 5 \times 5)$.
We consider four combinations of spatial dimensions. They are: $(3
\times 3, 3 \times 3)$, $(3 \times 3, 5 \times 5)$, $(5 \times 5, 3
\times 3)$ and $(5 \times 5, 5 \times 5)$.  Different filter sizes
result in different receptive field sizes of the FF-CNN. In turn, they
yield different features at the output of the conv layer. 

\noindent
{\bf Scheme 2) Subsets of derived features.} For each FF-CNN, we have the 
following feature set for each sample:
$$
{\bf F} = \{ {\bf F}_{conv1}, {\bf F}_{conv2} \},
$$
where ${\bf F}_{conv1}$ and ${\bf F}_{conv2}$ represent the features
obtained from the first and the second conv layers, respectively.  We
select a subset ${\bf V_i}$ from ${\bf F}$. There are many possible
selection choices.  We test the following three selection rules in the
experiment. 
\begin{enumerate}
\item For each channel in ${\bf F}_{conv1}$, select $\lambda_0 \times
W_1 \times H_1$ features randomly, where $W_1 \times H_1$ are the
spatial dimensions of the first conv layer and $\lambda_0 < 1$.  Then,
apply C-PCA to reduce the feature dimension to $K_1$.  Finally, randomly
select $\lambda_1 K_1$ features from $K_1$ features, where $\lambda_1 <
1$. 
\item Apply C-PCA to ${\bf F}_{conv2}$ to generate $K_2$
features and select $\lambda_2 K_2$ features randomly, where
$\lambda_2 < 1$.
\item Conduct checkerboard partitioning of ${\bf F}_{conv1}$ in the
spatial dimension. Then, apply the C-PCA to each part and generate two
feature subsets. 
\end{enumerate}
We generate one decision vector using each feature subset. 

\noindent
{\bf Scheme 3) Flexible input image forms.} We adopt different image
input forms to increase diversity. For example, we use various color
models to represent color images \cite{ibraheem2012understanding}. Here,
we use the RGB, YCbCr and Lab color spaces as different input forms to
an FF-CNN.  We also apply Laws filter bank of size $3 \times 3$
\cite{laws1980rapid} to input images to capture their different spectral
characteristics. The final decision vector is obtained by combining
FF-CNNs using different input representations. 

\subsection{Separation of Easy and Hard Examples}\label{subsec:separation}

It is desired to separate hard examples from easy ones in the
decision-making process. This is accomplished by computing the
confidence score of each decision. It is determined by two factors: 1)
the final decision vector of the ensemble classifier and 2) the
prediction results of all base classifiers. Intuitively speaking, a
decision is more confident if the maximum probability in the ensemble
decision vector is larger or more base classifiers in an ensemble agree
with each other. We define two confidence scores for an input image
${\bf X}_i$, where $i$ is the image index, as
\begin{equation}
CS1_i = max(P_{final}({\bf y}|{\bf X_i})), \quad CS2_i = N_{i}/ N_{all}.
\end{equation}
where ${\bf X}_i$, $CS1_i$, and $CS2_i$ denote the input data and two
confidence scores, respectively, ${\bf y}$ is the decision vector of the
ensemble, $N_{i}$ is the number of base classifiers producing the
majority class label for input ${\bf X_i}$ and $N_{all}$ is the total
number of the base classifiers. We call an input image a hard sample if
$CS1_i < T_1$ and $CS2_i < T_2$, where $T_1$ and $T_2$ are two threshold
values.  After the separation of easy and hard examples, a new FF-CNN
ensemble targeting at the hard samples set, ${\bf X}_{hard}$, can be
trained to boost the classification performance of hard samples. 

\section{Experiments}\label{sec:experiment}

We conducted experiments on two popular datasets: MNIST
\cite{lecun1998gradient} and CIFAR-10 \cite{krizhevsky2009learning}. The
MNIST dataset contains gray-scale images of handwritten digits 0-9.  The
CIFAR-10 dataset has 10 classes of tiny images of size $32\times32$.  We
adopted the LeNet-5 architecture \cite{Lecun98gradient-basedlearning}
for the MNIST dataset. Since CIFAR-10 is a color image dataset, we set
the filter numbers of the first and the second conv layers and the first
and the second FC layers to 32, 64, 200 and 100, respectively, by
following \cite{kuo2018interpretable}. 

We applied C-PCA to the output of the second conv layer and reduced the
feature dimension of the second conv layer per channel from 25 to 20
(for MNIST) or 12 (for CIFAR-10).  We sometimes fed the responses from
the first conv layer to the FC layers directly to increase feature
diversity. When this happens, we set reduced feature dimension per
channel to 30 (MNIST) and 20 (CIFAR-10) of the first conv layer while
the original dimension per channel is $14\times14=196$.  

We adopted the Radial Basis Function (RBF) SVM classifier as the
ensemble classifier in all experiments.  We applied PCA to cascaded
decision vectors of base classifiers before the SVM classifier training.
The reduced feature dimension was determined by the correlation of
decision vectors of base classifiers in an ensemble. 

\subsection{Performance of Ensemble Systems}

To show the power of ensembles, we conducted experiments by taking
diversity schemes discussed in Sec. \ref{sec:methods} into account. 

\noindent
{\bf Scheme 1.} We compare the performance of BP-CNN, four FF-CNNs and
the ensemble of four FF-CNNs in Table \ref{table:accuracy_2}. The four
FF-CNNs differ in their filter sizes in two conv layers: 1) (5x5,5x5),
2) (3x3,5x5), 3) (5x5,3x3), and 4) (3x3,3x3). For MNIST, their filter
numbers are the same in all settings; namely, (6,16).  For CIFAR-10,
their filter numbers for RGB images are: 1) (32,64), 2) (24,64), 3)
(32,64), and 4) (24,48).  Their filter numbers for a single channel of
color images are: 1) (16,32), 2) (8,32), 3 (16,32), and 4) (8,24). The
classification accuracies of BP-CNN, four FF-CNNs and the ensemble are
listed from columns 1 to 6.  We see that the ensemble of four FF models
provides 4\% improvement than the best single FF model.  Different
filter sizes will directly affect the receptive field size of each conv
layer and induce different statistics of the input data. In this way, we
introduce diverse features into the ensemble system.  While the
performance gap between BP-CNN and the ensemble narrows down for MNIST,
the ensemble outperforms BP-CNN for CIFAR-10. 

\begin{table}[h!]
\normalsize
\centering
\caption{Comparison of classification accuracy (\%) of BP-CNN,
four FF-CNNs and their ensemble on MNIST and CIFAR-10.} \label{table:accuracy_2}
\begin{tabular}{|c|ccccc|c|} \hline
 & BP&FF-1  & FF-2 & FF-3 & FF-4 & Ens. \\ \hline
 Filter Size & (5,5)& (5,5) & (5,3) & (3,5) & (3,3) & - \\ \hline
 MINST &{\bf99.1} & 97.1 & 97.0 & 97.2 & 97.3 & 98.2\\ \hline
 CIFAR-10 &68.7& 63.7 & 65.3 & 64.2 & 65.9 & {\bf 69.9} \\ \hline
\end{tabular}
\end{table}

\noindent
{\bf Scheme 2.} We evaluate the FF-1 design with feature subset
diversity and set $\lambda_0$, $\lambda_1$ and $\lambda_2$ to 75\%.  We
show the performance in Table \ref{table:accuracy_3}, where the first to
the fifth columns correspond to selected feature subsets from the entire
${\bf F}_{conv2}$ (denoted by Conv2), two chosen by the third rule
(denoted by Conv1-1 and Conv1-2), one by the first rule (denoted by
Conv1-RD), and one by the second rule (denoted by Conv2-RD),
respectively, where "RD" denotes reduced dimension.  Then, we study the
performance of four ensemble methods: 1) the ensemble of Conv1, Conv1-1,
Conv1-2 (ED-1); 2) the ensemble of six Conv1-RD results (ED-2); 3) the
ensemble of twelve Conv2-RD results (ED-3); and 4) the ensemble of six
Conv1-RD and twelve Conv2-RD results (ED-4).  As compared with the
performance of FF-1 for CIFAR-10 which is 63.7\%, we see that ensembles
using the feature subset diversity boost its performance by a margin
ranging from 2.3 to 5.6\%.  It is worthwhile to point out that one can
combine three classifiers (one trained on ${\bf F}_{conv1}$ feature set
and two trained on ${\bf F}_{conv2}$ feature set) in the ED-1 ensemble.
It yields 68.7\% and 97.7\% accuracy on CIFAR-10 and MNIST,
respectively. This choice offers a simple and effective ensemble system.
We will adopt ED-1 to build a larger ensemble system by adding other
sources of diversity later. 

\begin{table*}[h!]
\normalsize
\centering
\caption{The testing classification accuracy (\%) on MNIST and CIFAR-10
using feature subset diversities.}\label{table:accuracy_3}
\begin{tabular}{|c|ccccc|cccc|} \hline
 & Conv2  & Conv1-1 & Conv1-2 & Conv1-RD & Conv2-RD & ED-1 & ED-2 & ED-3 & ED-4\\ \hline
 MINST & 97.1 & 95.4 & 95.3 & 96.8 & 95.2 & 97.7 & 97.6 & 97.2 & \bf98.0\\ \hline
 CIFAR-10 & 63.7 & 64.3 & 64.4 & 62.3 & 64.2 & 68.7 & 66.0 &  68.4 & \bf69.3 \\ \hline
\end{tabular}
\end{table*}

\noindent {\bf Scheme 3.} We conduct experiments by adopting different
inputs to the FF-1 architecture in Table \ref{table:accuracy_4}. We
apply nine Laws filters of size $3 \times 3$ \cite{pratt2007digital} to
gray-scale images and generate nine images that contain frequency
components in different subbands. For color images in CIFAR-10, we
represent the color information in three color spaces: RGB, YCbCr, and
Lab, where we treat three channels individually in the last two color
spaces. We observe 1.1\% and 5.9\% performance improvements on the MNIST
and CIFAR-10, respectively, by assembling various input representations.
This demonstrates the effectiveness of utilizing various input
representations as the diversity source in an ensemble. 

\begin{table*}[h!]
\centering
\footnotesize
\caption{The testing classification accuracy (\%) on MNIST and
CIFAR-10, where L1 to L9 denote the filtered maps by L3L3,
E3E3, S3S3, L3S3, S3L3, L3E3, E3L3, S3E3, and E3S3 Laws filters,
respectively. The last column indicates the ensemble results.} 
\label{table:accuracy_4}
\begin{tabular}{|c|cccc|ccccccccc|c|} \hline
  &RGB & Grey & YCbCr & Lab &L1  & L2 & L3 & L4 & L5 & L6 & L7 & L8 & L9 & ED\\ \hline
  MINST &- & 97.1 & - & - & 97.0 & 95.1 & 87.8 & 92.6 & 93.7 & 94.9 & 95.6 & 93.1 & 92.6 & \bf98.2\\ \hline
  CIFAR-10 & 63.7 &-& 54.0/41.4/41.1 & 53.2/40.0/41.0 & 50.6 & 44.8 & 44.5 & 46.3 & 48.3 & 44.9 & 47.6 & 43.0 & 45.8 & \bf69.6\\ \hline
\end{tabular}
\end{table*}

\begin{figure}[h!]
\centering
\includegraphics[width=0.6\linewidth]{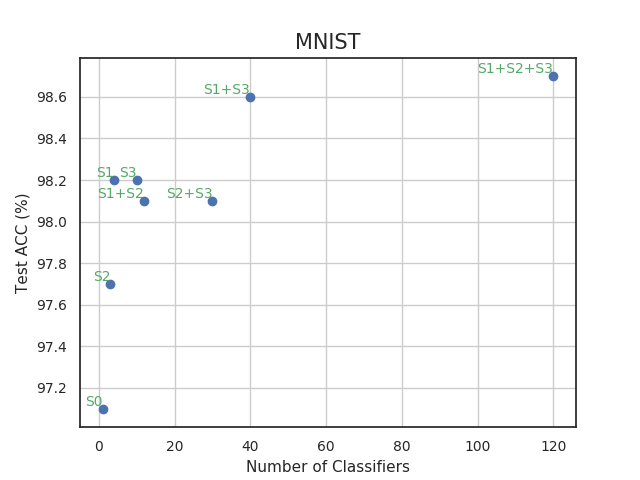}
\includegraphics[width=0.6\linewidth]{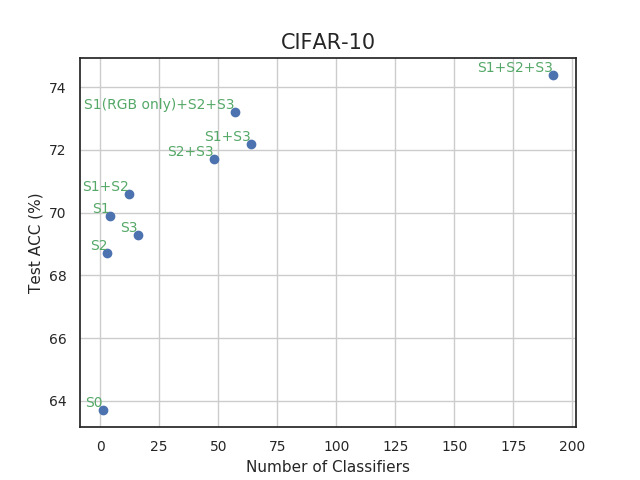}
\caption{The relation between test accuracy (\%) and the number of
FF-CNNs in the ensemble, where three diversity sources are indicated as
S1 (Scheme 1), S2 (Scheme 2) and S3 (Scheme 3).} \label{fig:acc_overall}
\end{figure}

We can fuse three diversity types in an ensemble to boost the
performance.  The relation between test accuracy and ensemble complexity
is shown in Fig.  \ref{fig:acc_overall}. In general, the ensemble of
more classifiers gives better performance. So far, the best performance
achieved on MNIST and CIFAR-10 are 98.7\% and 74.2\% in terms of test
accuracy. As compared with the single BP-CNN reported in Table
\ref{table:accuracy_2}, the best ensemble result is 5.5\% higher on
CIFAR but 0.4\% lower on MNIST.  We can push the performance higher by
separating easy and hard examples based on the scheme described in Sec.
\ref{subsec:separation}. 

\subsection{Separation of Easy and Hard Examples}

By following the discussion in Sec.  \ref{subsec:separation}, we set
$T_1=0.98$ and $T_2=0.7$ for the MNIST dataset and $T_1=0.97$ and
$T_2=0.65$ for the CIFAR-10 dataset. The results are reported in Table
\ref{table:accuracy_5}. For the set of hard examples, the new ensemble
system trained on this set provides 5.6\% and 2.6\% improvements in test
accuracy for MNIST and CIFAR-10, respectively. More hard samples are
classified correctly in this setting.  Overall, the ensemble method 
with easy/hard example separation achieves test accuracies of 99.3\% 
and 76.2\% on the entire MNIST and CIFAR-10 datasets, respectively. 
It outperforms the best results obtained earlier as shown in Table 
\ref{table:accuracy_2}.

\begin{table}[h!]
\centering
\footnotesize
\caption{The classification accuracy (\%) on MNIST and CIFAR-10
datasets, where the first, second and fourth columns indicate results of
the original ensemble system evaluating on the easy, hard and the entire
sample sets accordingly. The third and fifth columns present the results
of the new ensemble system evaluating on the hard set and the entire
set, respectively.}\label{table:accuracy_5}
\begin{tabular}{|c|c|ccccc|} \hline
\multicolumn{2}{|c|}{}  &  Easy  & Hard & Hard+ & FF  & FF$^{+}$\\ \hline
\multirow{ 2}{*}{{MNIST}} 
&Train & 99.9 & 90.0 &98.2& 98.9 & \bf99.7 \\ 
&Test  & 99.9 & 88.0 &93.6& 98.7 & \bf99.3 \\ \hline
\multirow{ 2}{*}{Cifar-10} 
&Train & 99.9 & 73.5 &82.3& 80.1 &\bf87.2 \\ 
&Test  & 98.2 & 66.2 &68.8& 74.2 & \bf76.2 \\ \hline
\end{tabular}
\end{table}

\subsection{Discussion}

To better understand the diversity among different FF-designed CNNs, we
evaluate the correlation among the output of different classifiers using
two diversity measures: Yule’s Q-statistic and entropy measure
\cite{cunningham2000diversity}. These measures are built on the
correct/incorrect decision. The lower Q-statistic (or the higher entropy
measure) indicates a higher diversity degree among base classifiers. The
average measures among different diversity sources are reported in Table
\ref{table:accuracy_6}. The best diversity measures are achieved by
combining all base classifiers, leading to a large amount of performance
improvement.  This is consistent with classification accuracy assessment
as shown in Fig. \ref{fig:acc_overall}. 

\begin{table}[h!]
\normalsize
\centering
\caption{Diversity measures on CIFAR-10. Three types of diversity
sources are evaluated in the first to third columns, respectively, and
the last column reports the measure on all base classifiers in an
ensemble.}\label{table:accuracy_6}
\begin{tabular}{|c|ccc|c|} \hline
            & S1 & S2 & S3 & ALL\\ \hline
Q-statistic & 0.88 & 0.89 & 0.66 & 0.61 \\ \hline
Entropy measure & 0.21 & 0.24 & 0.47 & 0.49 \\ \hline
\end{tabular}
\end{table}


\section{Conclusion}\label{sec:conclusion}

We proposed an ensemble method that is built on multiple FF-CNNs of
diversity. We see a significant improvement in test accuracy for the
MNIST and the CIFAR-10 datasets.  As future extensions, we would like to
apply the ensemble method to the more challenging datasets with more
object classes or/and larger image size, such as the CIFAR-100 and the
ImageNet. Also, it will be interesting to develop a weakly supervised
system based on the ensemble of multiple FF-CNNs. 

\bibliographystyle{IEEEbib}
\bibliography{icme2019template}

\end{document}